# A Real Time Facial Expression Classification System Using Local Binary Patterns

S L Happy, Anjith George, and Aurobinda Routray
Department of Electrical Engineering,
IIT Kharagpur, India

*Abstract*— **Facial expression analysis is one of the popular fields of research in human computer interaction (HCI). It has several applications in next generation user interfaces, human emotion analysis, behavior and cognitive modeling. In this paper, a facial expression classification algorithm is proposed which uses Haar classifier for face detection purpose, Local Binary Patterns (LBP) histogram of different block sizes of a face image as feature vectors and classifies various facial expressions using Principal Component Analysis (PCA). The algorithm is implemented in real time for expression classification since the computational complexity of the algorithm is small. A customizable approach is proposed for facial expression analysis, since the various expressions and intensity of expressions vary from person to person. The system uses grayscale frontal face images of a person to classify six basic emotions namely happiness, sadness, disgust, fear, surprise and anger.**

*Index Terms*-- **Facial expression analysis, Haar classifier, Principal Component Analysis(PCA), multi-block Local Binary Patterns(LBP) features.**

## I. Introduction

Intelligent Human-Computer Interaction (HCI) aims at efficient interaction between human and machines using body gestures, eye gaze, speech, facial expressions, cognitive modeling etc. Perceptual recognition, machine learning, cognitive modeling and affective computing are the emerging trends of interest in recent years. Facial expressions can be used as an efficient way of emotion detection, thus facilitating HCI. Detection and classification of facial expressions can be used as a natural way of interaction between man and machine. However facial expressions and its intensity vary from person to person and also varies along with age, gender, size and shape of face, and further, even the expressions of the same person do not remain constant with time. Hence, designing a generalized method for facial expression analysis is a very difficult task. However, customized algorithms can be used to analyze facial expressions of a single person accurately.

Facial expression analysis gained interest with recent advancements in face detection, face tracking, and face recognition techniques. B. Fasel and J. Luettin [1] described different types of approaches for facial expression analysis like model based methods, holistic methods, local methods and motion extraction methods. They have also considered different issues of difficulties involved in expression recognition such as pose, illumination and occlusions. Ekman and Keltner developed Facial Action Coding System (FACS) [2] which describes facial expressions in terms of Action Units (AUs). FACS consists of 46 AUs which, basically, are related to facial muscle movements. A Principal Component Analysis (PCA) based method for classifying expressions is given by Juanjuan *et al.* [3]. In [4] , Cohen *et al.* have used True Augmented Naive Bayes(TAN) classifiers to find dependencies among different facial expressions and Hidden Markov Model(HMM) to automatically segment and recognize them. Wang *et al.* [5] proposed real time facial expression recognition using AdaBoost technique in which they have first detected face using Haar features, then recognize six basic expressions from the features extracted from the detected face region. In [6] PCA is applied to local regions in face image like lips and eye images and feed these features to a Hierarchical Radial Basis Function Network (HRBFN) to classify the facial expressions. However, for this purpose lips and eye regions need to be detected first and improper detection would degrade the performance. Using geometric, physical and motion-based dynamic model of facial features, Essa and Pentland [7] accurately estimate facial motions in video sequences. However, classification of emotions through spatial information from static images is very complicated. Anderson and McOwan [8] have used optical flow and Support Vector Machine (SVM) to classify emotions in real time. They have achieved 90% classification accuracy when the confusing expressions are ignored. A Local Global Graph (LGG) [9] approach has been reported which can recognize facial expressions irrespective of illumination conditions and complexity of background. Active Appearance Model (AAM) is used in [10] to extract facial features which are further classified using SVM. However, for real time implementation of such algorithms, time complexity plays a crucial role which usually traded off with accuracy.

In this paper, we have proposed a real time customized algorithm that can classify facial expressions of a person into the basic emotions namely anger, fear, sadness, happiness, disgust and surprise. To reduce computational complexity, certain selected feature points have been used rather than the entire image itself for classification purpose. PCA based dimensionality reduction is used owing to its better performance for face images [16] in illumination variation and occlusion. Being a local feature descriptor, Local Binary Patterns (LBP) features take care of minor changes of facial expressions for different emotions. After extracting local features from face images using LBP, different expressions are classified to different classes of emotions.

The paper is organized as follows. Section II describes the overview of the system. Section III describes the algorithms used for facial expression classification. Section IV explains the results obtained, followed by conclusion and future work in Section V.

This work is financially supported by Department of Information Technology, Government of India.

## II. SYSTEM OVERVIEW

A customized real time implementable algorithm for facial expression classification is proposed in this paper. The system is trained using supervised learning approach in which it takes images of different facial expressions in the training phase along with ground truth values as inputs. Face detection and feature extraction are carried out from face images and using the feature vector, a PCA based classifier is constructed for each class of emotions. A personalized custom training for each user is proposed to reduce the false detections. Once the system is calibrated for a particular user, the training data can be loaded each time and the facial expression classification can be done in real time.

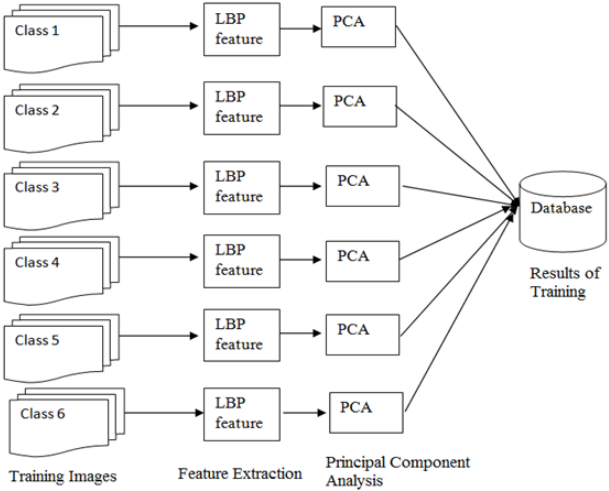

Fig. 1. Flowchart for training phase

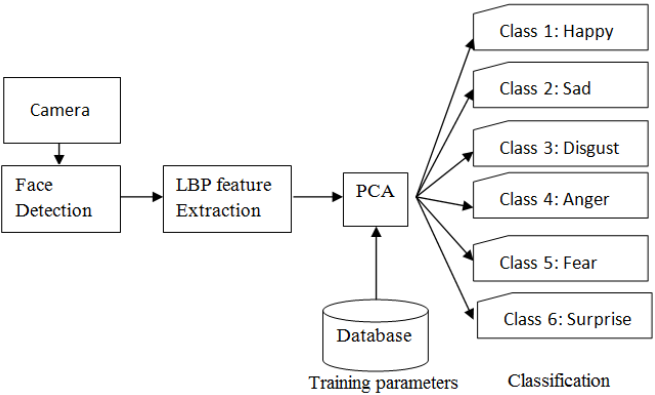

Fig. 2. Flowchart for testing phase

### A. Training stage

Fig. 1 describes the stages of training phase to construct the customized database. Six classes belonging to six basic expressions are considered to which the training samples are to be mapped. Face detection is carried out in the training dataset using Haar classifiers. Then, block LBP histogram features are obtained from the face images. PCA is used to reduce the dimensionality of data by finding the eigen directions in which same class features are more dispersed. The training is carried out for all six classes of facial expressions and the principal eigenvectors for each class are stored in a Comma Separated Value (CSV) file.

### B. Detection stage

The detection phase consists of identifying the face from the image captured from camera followed by calculation of block LBP histogram features and classification of feature vector using PCA for each class. The principal eigen vectors stored in the database for each class are invoked. The block LBP histogram features are calculated from the detected face image and projected into different eigen directions of each class. The reconstruction error for each class is calculated and the face is classified to the class corresponding to the minimum reconstruction error. Figure 2 describes the different stages of detection phase.

## III. METHODOLOGY

The proposed algorithm mainly consists of three steps face detection, feature extraction, and facial expression classification. Face detection is carried out using Haar classifier based method.

### A. Face detection

Haar classifier based method is chosen for face detection owing to its high detection accuracy and real time performance [11]. Haar like features encodes the difference in average intensity in the different parts of the image and consists of black and white connected rectangles in which the value of the feature is the difference of sum of pixel values in black and white regions. Haar like features can be found out at different scales and positions which are robust for detection purpose.

The computational speed of the feature calculation is increased with the use of Integral image [11]. The number of Haar features available from a small image is very large. The optimal set of features to be used and their corresponding thresholds to be used for classification is obtained from AdaBoost algorithm [5], [11].

### B. Feature Extraction

Selection of the feature vector is the most important part in a pattern classification problem. The inherent problems related to image classification include the scale, pose, translation and variations in illumination level. The feature vector extracted should be invariant to these problems to achieve better accuracy in classification. In facial expression classification and face recognition Gabor features are very popular due to its high discriminative power. But the computational complexity and memory requirement make it less suitable for a real time implementation [12]. LBP based feature extraction method is used owing to its excellent light invariance property and low computational complexity [13]. The neighbourhood values are thresholded by the center value and the result is treated as a binary number. In this way, it encodes the neighbourhood information very efficiently.

Fig. 3 shows the LBP feature calculation procedure. The values in 3x3 neighborhoods are thresholded and their binary weighted summation gives the feature value at that location.

$$LBP = \sum_{n=0}^{7} s(i_n - i_c) 2^n \quad (1)$$

Where, $i_c$ is the central pixel value and

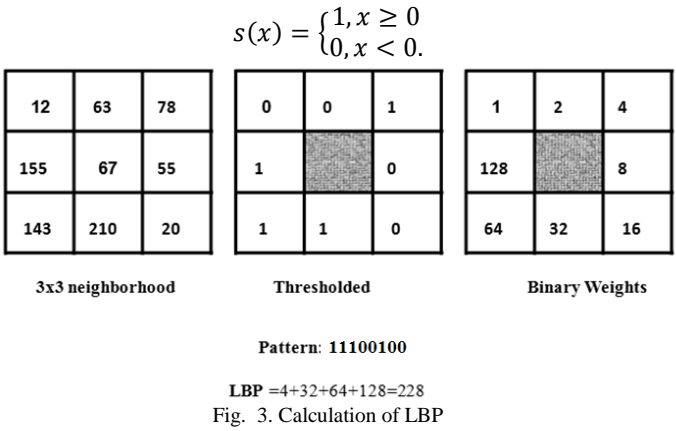

Fig. 3. Calculation of LBP

In LBP histogram, the LBP coefficients are collected in a histogram, but it results in the loss of spatial information. The spatial information is retained by dividing the image into several small blocks, then the LBP histogram of each small region is found, and finally all these features are concatenated to give a global descriptor. Such a global descriptor contains information in three levels global features, region features and pixel level features [14].

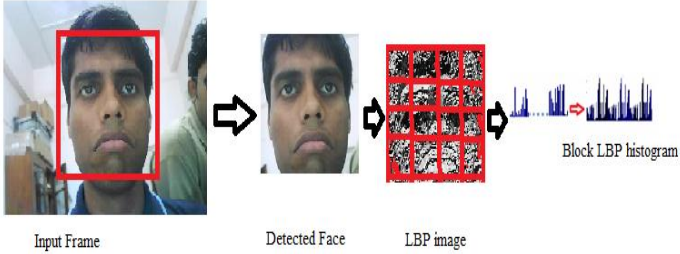

Fig. 4. Calculation of block LBP histogram

In multi block LBP feature extraction approach, the image is divided into small subsections and the LBP histogram for each block is calculated. This feature is calculated in multiple levels. Finally the histograms are concatenated to form the global feature vector. Figure 4 explains the idea of block LBP histogram technique.

*C. Facial Expression Classification*

The dimensionality of data obtained from the feature extraction method is very high. Dimensionality reduction is done with PCA to improve the speed of computation in real time.

PCA allows us to compute a linear transformation that maps data from a high dimensional space to a lower dimensional subspace. PCA uses an orthogonal transformation to convert a set of observations of possibly correlated variables into a set of values of uncorrelated variables called principal components [15]. Here each expression is considered as a class. The training and detection algorithms are given below:

TABLE I. TRAINING ALGORITHM

1. Detect face from the training image using Haar classifier and resize detected face image to $N \times M$ resolution
2. Preprocess the face image to remove noise
3. For each class(expression) obtain the feature vectors $\Gamma_{j,1}, \Gamma_{j,2}, \ldots, \Gamma_{j,P}$ ($j^{th}$ class) of dimension $\left(\frac{N}{n} * \frac{M}{m} * b, 1\right)$ each
   a. Divide the face image to sub-images of resolution $n \times m$, find the LBP values and calculate $b$-bin histogram for each block
   b. Concatenate the histograms of each block to get the feature vector($\Gamma_{j,i}$) of size $\left(\left(\frac{N}{n} * \frac{M}{m}\right) * b, 1\right)$
4. Compute the mean feature vector of individual class

$$\psi_j = \frac{1}{P}\sum_{i=1}^{P} \Gamma_{j,i} \quad (j = 1,2,\ldots,6) \quad (2)$$

5. Subtract the mean feature vector from each feature vector $\Gamma_{j,i}$

$$\phi_{j,i} = \Gamma_{j,i} - \psi_j \quad (j = 1,2\ldots,6) \quad (3)$$

6. Estimated covariance matrix $C$ for each class, given by

$$C_j = \frac{1}{P}\sum_{i=1}^{P} \phi_{j,i}\phi_{j,i}^T = A_j A_j^T \quad (j = 1,2,\ldots,6) \quad (4)$$

Where $A_j = [\phi_{j,1}, \phi_{j,2}\ldots \phi_{j,P}]$ of dimension $\left(\frac{N}{n} * \frac{M}{m} * b\right) \times P$ which is very large.
Compute $A_j^T A_j$ ($P \times P$) instead as $P \ll \left(\frac{N}{n} * \frac{M}{m}\right) * b$

7. Compute the eigenvectors $v_{j,i}$ of $A_j^T A_j$ using the equation

$$\sigma_{j,i} u_{j,i} = A_j v_{j,i} \quad (j = 1,2\ldots,6) \quad (5)$$

Eigenvectors $u_{j,i}$ of $A_j A_j^T$ are obtained.
8. Keep only $K$ eigenvectors corresponding to the $K$ largest Eigen values from each class (suppose $U_j = [u_{j,1}, u_{j,2}, \ldots, u_{j,k}]$)
9. Normalize the $K$ eigenvectors of each class.

TABLE II. ALGORITHM USED TO IMPLEMENT PCA FOR FACIAL EXPRESSION DETECTION

1. Detect face with Haar classifier algorithm and resize face image to $N \times M$ resolution
2. Preprocess the face image to remove noise
3. Find feature vector ($\Gamma$) for the resized face using similar methods as used in training phase (Step 3).
4. Subtract the mean feature vector of each class from $\Gamma$

$$\phi_j = \Gamma - \psi_j \quad (j = 1, 2..6) \quad (6)$$

5. Project the normalized test image onto the eigen directions of each class and obtain weight vector

$$W = [w_{j,1}, w_{j,2}, \ldots, w_{j,k}] = U_j^T \phi_j \quad (j = 1,2,\ldots,6) \quad (7)$$

6. Compute

$$\widehat{\phi_j} = \sum_{1}^{k} w_{j,i}\, u_{j,i} \quad (j = 1,2,\ldots,6) \quad (8)$$

7. Compute error

$$e_j = \|\phi - \widehat{\phi_j}\| \quad (j = 1,2,\ldots,6) \quad (9)$$

The image is classified to the training set, to which it is closest (when the reconstruction error ($e_j$) is minimum).

IV. EXPERIMENTAL RESULTS

Facial expressions for different emotions were captured for a particular subject since the algorithm is customized for an individual. For each subject a training set consisting of 540 images with 90 images for each expression and a testing set of 900 images with 150 from each emotion was created.

## A. Training phase

First the face is detected from training set images by Haar classifier based on Haar like features. The false detections are rejected manually. The detected face region is cropped and resized to a size of $40 \times 40$ pixels. Bilateral filtering is carried out on the image to remove noise. The preprocessed face image is used to extract features of facial expression using LBP algorithm. Block LBP approach is used in which the face image is subdivided to many sub-images and LBP features of each block are calculated. The bin histogram of each sub-image LBP features is determined and hence by concatenation of the block bin histograms the training vector was created. The feature vector is used as input to PCA and the orthogonal vectors for each class of emotional facial expression are generated. A few numbers of orthogonal vectors, 40 in our case, along maximum dispersion of data were chosen for each expression and stored in the database.

TABLE III. AVERAGE CONFUSION MATRIX

|  | Anger | Disgust | Fear | Happiness | Sad | Surprize |
|---|---|---|---|---|---|---|
| Anger | 0.966 | 0.000 | 0.000 | 0.000 | 0.027 | 0.007 |
| Disgust | 0.000 | 1.000 | 0.000 | 0.000 | 0.000 | 0.000 |
| Fear | 0.027 | 0.000 | 0.920 | 0.007 | 0.046 | 0.000 |
| Happiness | 0.000 | 0.000 | 0.000 | 1.000 | 0.000 | 0.000 |
| Sad | 0.020 | 0.007 | 0.000 | 0.000 | 0.973 | 0.000 |
| Surprize | 0.000 | 0.000 | 0.000 | 0.007 | 0.000 | 0.993 |

## B. Testing phase

In the testing phase, the face is first detected using Haar classifier and resized to $40 \times 40$ pixels, the same resolution used for training stage. Then the bin histogram of block LBP for face image is calculated as given in training phase. The feature vector is projected along the eigen direction of each class and weight vector in each eigen direction is obtained. Using the weight vectors, the feature vector is reconstructed back and reconstruction error is calculated. Performing such projection of feature vector to eigen direction of each class of expression, distance of testing image from each of the classes is calculated. The testing face image is classified to the class of expression according to minimum distance criterion.

Experiments with block LBP features of different block sizes are carried out and it is found that using cascaded multi-block LBP features classification becomes more accurate. i.e. when block LBP histogram features for two different block sizes are concatenated, the accuracy increases. From experiment it is found that optimum accuracy with less computation time is obtained by using block sizes of $6 \times 6$ and $8 \times 10$ pixels for calculation of LBP histogram features. However, higher dimension of feature vector increases computational time achieving higher accuracy.

Real time facial expression classification using two level block LBP histogram features has been implemented as shown in Fig. 5. After face detection, LBP histogram vectors for block sizes $6 \times 6$ and $8 \times 10$ pixels are carried out. The numbers of histogram bins used are 8 and 16 respectively. The bar graph shown in Fig. 5 represents the similarity of the input face image to different class of expressions. The algorithm is implemented in a PC with Intel Pentium i5 CPU and speed of 3.2 GHz using MATLAB and a speed of 12 fps is achieved.

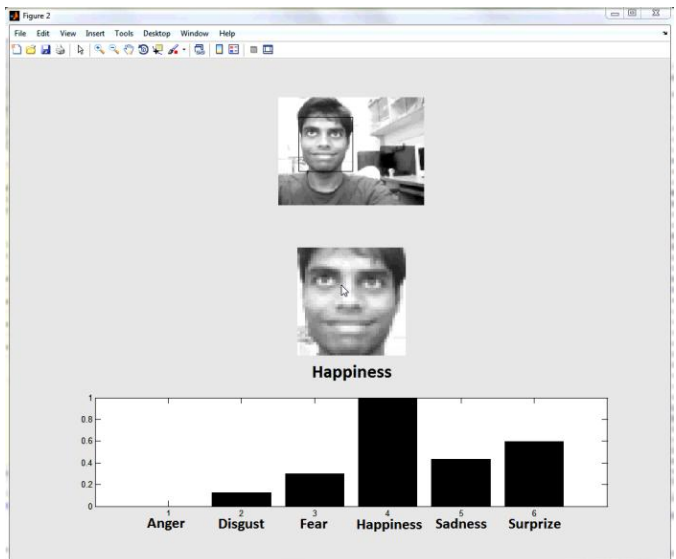

Fig. 5. Real time detection of face and facial expression classification

The algorithm was tested on the testing set and the recognition accuracy obtained for the six basic facial expressions was above 97%. Figure 6 shows some of the expressions which are correctly classified. The average confusion matrix for the dataset considered is shown in Table III. It is observed that angry and sad expressions have similar features and hence misclassified often to one another.

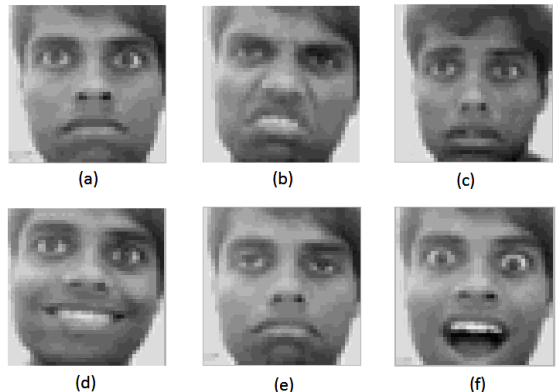

Fig. 6. Some correctly classified expressions (a) Anger, (b) Disgust, (c) Fear, (d) Happiness, (e) Sadness, (f) Surprize

## V. CONCLUSION

A real time customized algorithm for facial expression classification has been developed. Experimentally, the size of the block for LBP feature extraction is chosen for higher recognition accuracy. The testing results indicate that by using LBP features facial expressions recognition accuracy is more than 97%. The block LBP histogram features extract local as well as global features of face image resulting higher accuracy. Use of high resolution image along with more cascaded block LBP feature can increase classification accuracy. However, practically it is difficult to implement in real time due to associated time complexity.

The proposed methodology is limited to classify frontal image only. However, rotation of face or occlusions degrades the performance of the system. In the future, the research will be extended to 3-D face modeling using multiple cameras to improve the proposed facial expression recognition system.

ACKNOWLEDGMENT

The funds received from the Department of Information and Technology for conducting the research work is gratefully acknowledged.